# CGEarthEye:A High-Resolution Remote Sensing Vision Foundation Model Based on the Jilin-1 Satellite Constellation

Zhiwei Yi[1,2], Xin Cheng[1], Jingyu Ma[1], Ruifei Zhu[1*], Junwei Tian[1], Yuanxiu Zhou[1], Xinge Zhao[1], HongzheLi[1]

(1.Chang Guang Satellite Technology Co., Ltd, China；2.College of Electronic Science and Engineering, JiLin University, China)

**Abstract**：Deep learning methods have significantly advanced the development of intelligent rinterpretation in remote sensing (RS), with foundational model research based on large-scale pre-training paradigms rapidly reshaping various domains of Earth Observation (EO). However, compared to the open accessibility and high spatiotemporal coverage of medium-resolution data, the limited acquisition channels for ultra-high-resolution optical RS imagery have constrained the progress of high-resolution remote sensing vision foundation models (RSVFM). As the world's largest sub-meter-level commercial RS satellite constellation, the Jilin-1 constellation possesses abundant sub-meter-level image resources. This study proposes CGEarthEye, a RSVFM framework specifically designed for Jilin-1 satellite characteristics, comprising five backbones with different parameter scales with totaling 2.1 billion parameters. To enhance the representational capacity of the foundation model, we developed JLSSD, the first 15-million-scale multi-temporal self-supervised learning (SSL) dataset featuring global coverage with quarterly temporal sampling within a single year, constructed through multi-level representation clustering and sampling strategies. The framework integrates seasonal contrast, augmentation-based contrast, and masked patch token contrastive strategies for pre-training. Comprehensive evaluations across 10 benchmark datasets covering four typical RS tasks demonstrate that the CGEarthEye consistently achieves state-of-the-art (SOTA) performance. Further analysis reveals CGEarthEye's superior characteristics in feature visualization, model convergence, parameter efficiency, and practical mapping applications. This study anticipates that the exceptional representation capabilities of CGEarthEye will facilitate broader and more efficient applications of Jilin-1 data in traditional EO application. The code and pre-trained model weights will be released at: https://github.com/1921134176/CGEarthEye.

## 1 Introduction

The Jilin-1 satellite constellation, currently the world's largest sub-meter-level commercial remote sensing (RS) satellite constellation operated by Chang Guang Satellite Technology Co., Ltd. (CGST), demonstrates exceptional observation capabilities with: 6 global coverage cycles annually, full coverage of China every 15 days and 38-40 daily revisits to any global location. Its operational capabilities have been extensively validated in strategic applications spanning national security surveillance, precision agriculture monitoring, ecological environment assessment, and smart city planning [1-6]. Confronted with the high-frequency, massive data streams from the Jilin-1 satellite constellation, conventional RS interpretation approaches relying on machine learning and manual analysis are increasingly inadequate for contemporary operational demands [7, 8]. The development of vision foundational models leveraging massive multi-temporal Jilin-1 satellite imagery to support diverse interpretation tasks poses a significant scientific challenge.

Deep learning has significantly advanced RS image interpretation. Computer vision models like ResNet [9]、 DeepLabV3 [10]、HRNet [11]、ConvNeXt [12] now enable superior performance in specific tasks [13, 14]. All the aforementioned methods adopt a transfer learning approach by applying pre-trained model weights from the computer vision domain to the remote sensing domain. However, due to the significant domain gap between natural images in computer vision and remote sensing imagery, the models still heavily rely on high-quality annotated remote sensing data and exhibit limited generalization performance [15-17].

To address these challenges, it is critical developing remote sensing vision fundation model (RSVFM) with enhanced image feature extraction [18-21]. The remote sensing community has long grappled with limited-scale annotated datasets, creating a critical bottleneck for advancing interpretation research [22-24]. Current remote sensing datasets, fMoW [25] and BigEarthNet [26, 27] with respective sizes of 132,716 and 590,326 annotated scenes, remain orders of magnitude smaller than natural image benchmarks like ImageNet-1K [28]. Long et al. (2021) bridged this gap by introducing the MillionAID dataset with 1,000,848 samples, the first remote sensing benchmark comparable in scale to ImageNet-1K, has catalyzed a paradigm shift in supervised pre-training methodologies [20, 29]. However, supervised pre-training exhibits inherent limitations in geospatial contexts: The annotation process for remote sensing imagery demands specialized domain expertise and intensive manual annotation, making it suboptimal for developing foundational geospatial models [30-33].

Constructing large-scale remote sensing annotated datasets faces challenges such as high annotation complexity and high costs. In light of this situation, how to effectively mine the potential value of unlabeled data has become a key breakthrough in building robust and generalizable RS foundation models [34]. Self-supervised learning (SSL) have demonstrated unique advantages. They are capable of extracting feature representations from vast amounts of unlabeled images [35-37], providing an innovative pathway to break through the dependence on labeled data. SSL are generally divided into two categories, contrastive learning [38, 39] and generative learning [40-43]. Contrastive learning drives the aggregation of features of similar samples and increases the distance between dissimilar samples by setting up proxy tasks. In the RS field, scholars often integrate geographical coordinate metadata [44-46] and temporal features [47, 48] to construct contrastive pre-training tasks. However, the model design and the data preparation of such tasks pose significant engineering challenges, and existing studies have primarily focused on medium-resolution satellite imagery, such as the Sentinel series. In comparison, the generative learning, such as masked image modeling (MIM), enhances the model's representation ability through an image reconstruction mechanism, and its efficiency has been verified in several RS pre-training studies.[18, 49-53]. Emerging research reveals that hybrid pre-training frameworks integrating discriminative and generative paradigms synergistically enhance feature representation capabilities [54-57]. In conclusion, Large RSVFM trained via SSL on massive imagery show superior accuracy and generalization.

While current RSFM are increasingly integrating diverse data sources and pretraining techniques, their development remains uneven. Benefiting from the open-access policy and high-frequency global coverage of the Sentinel satellite series, medium-resolution multispectral SSL datasets centered on Sentinel imagery have rapidly advanced [26, 27]. For instance, Manas et al., 2021 constructed a large-scale multi-temporal Sentinel-2 multispectral SSL dataset, employing seasonal contrastive pretext tasks to develop remote sensing foundational models. The European Space Agency (ESA) has established MajorTOM-Core, the largest publicly available Sentinel-2 imagery dataset to date. Under the MajorTOM framework, ESA further developed and released image embeddings datasets using open-source vision fundation models, driving advancements in VFM [58]. SkySense leverages a globally-scoped, self-curated dataset comprising long-term temporal Sentinel-2 multispectral and Sentinel-1 SAR observations to construct multimodal remote sensing foundation models through geo-prototypical representation and temporal characterization modeling [57]. However, existing research is constrained by the uncontrollability and scarcity of high-resolution data, which limits the spatiotemporal coverage of data and the development of pre-training algorithms. Most studies exclusively utilize MillionAID as the primary data source. Although it incorporates global sampling of Google Earth imagery, its spatial coverage and temporal span remain constrained [59]. This limitation partially impedes the advancement of high-resolution RFVFNs, since the lack of controlled data quality and diversity ultimately degrades model performance, which is critical for producing discriminative feature representations.

To address these challenges, this study leverages the autonomous and scalable massive database of Jilin-1

satellites. Through multi-stage representation clustering and adaptive sampling strategies, we constructed Jilin-1 Self-supervised Seasonal Dataset (JLSSD). To the best of our knowledge, JLSSD is the first large-scale remote sensing self-supervised dataset featuring global coverage, multi-seasonal observations within a single calendar year, and submeter-scale spatial resolution. Based on JLSSD, we proposed a multi-scale Contrastive learning framework integrating three synergistic tasks, augmentation-aware contrastive learning 、 seasonal alignment contrastive learning and masked patch token contrastive. This framework was employed to pre-train Vision Transformer (ViT) architectures, yielding the Jilin-1 Remote Sensing Visual Foundation Model Series (CGEarthEye). Extensive evaluations across 10 high-resolution benchmarks covering four critical Earth observation tasks (e.g., land cover classification, change detection, object recognition, and semantic segmentation) demonstrate that CGEarthEye achieves state-of-the-art (SOTA) performance in all scenarios. Furthermore, practical deployment tests utilizing Jilin-1 satellite data and real-world operational workflows confirm that CGEarthEye consistently outperforms previous compact models in industrial applications, while maintaining superior generalization capability.

In summary, the contributions of this study are threefold,

(1) We propose CGEarthEye, a RSVFM specifically designed for the Jilin-1 constellation, currently the world's largest commercial sub-meter remote sensing satellite system. The framework incorporates five backbone variants with 2.1 billion total parameters, adaptable to four downstream tasks including scene classification, object detection, semantic segmentation, and change detection.

(2) Through a clustering and spatiotemporal sampling strategy, we established JLSSD, the first 15-million-scale SSL dataset featuring global coverage and quarterly temporal sampling within a single year at 2023. By synergistically integrating augmentation-aware contrastive learning, seasonal contrastive alignment, and masked patch token contrastive learning, the framework significantly enhances feature representation learning for high-resolution remote sensing data.

(3) Extensive evaluations across 10 high-resolution benchmarks demonstrate that CGEarthEye achieves SOTA performance on all tested EO tasks. Notably, under frozen backbone settings, CGEarthEye outperforms existing remote sensing foundation models in both accuracy and generalization capability.

## 2 CGEarthEye

This section systematically presents the three core technical components of the CGEarthEye framework, SSL dataset, foundation model architecture and pre-training algorithm.

### 2.1 JLSSD

Leveraging the Jilin-1 constellation's extensive historical data, we construct JLSSD——a large-scale, globally covered, high-resolution SSL dataset, using multi-dimensional representation clustering and adaptive sampling strategie. The sampling process begins with partitioning the globe into 1 km × 1 km grid cells, followed by attribute stratification using ESA WorldCover land cover classification data, global Digital Elevation Model (DEM) data, and administrative boundary data. For an individual grid cell G, three key attributes are extracted. For an individual grid cell G, three key attributes are extracted, the land cover category $c$, elevation bin $i$ and administrative region $r$. The land cover category $c$ corresponds to the dominant land cover class within G derived from ESA WorldCover data, categorized into seven types including forest, grassland, cropland, water body, wetland, built-up area, and others. The elevation bin $i$ is assigned to one of 24 discrete segments generated by dividing the global elevation range (-2000 m to 10,000 m) into 500 m intervals. The administrative region $r$ corresponds to the jurisdictional unit of the grid cell: county-level divisions are adopted for Chinese territories, while country-level divisions are applied to non-Chinese regions in this study. Based on the defined attributes, let $S_{c\_i\_r}$ denotes the grid set characterized by the land cover category $c$, elevation bin $i$ and administrative region $r$. For each $S_{c\_i\_r}$ set, a randomized sampling process is performed, with the sampling frequency calculated as follows.

$$M_{c\_i\_r\_sample} = M \times W_c \times \frac{M_{c\_i\_r}}{M_c}$$

where $M_{c\_i\_r\_sample}$ denotes the number of sampled grid cells within subset $S_{c\_i\_r}$, $M_{c\_i\_r}$ denotes the total number of the $S_{c\_i\_r}$, $M_c$ is the number of global grid cells belonging to land cover category c.

Based on the aforementioned clustering and sampling rules, we divide the Chinese and non-Chinese regions into two distinct grid populations for independent sampling. For the Chinese region, quarterly mosaics from 2023 serve as the data source, while annual mosaics from 2023 are used to extract sampled grids for non-Chinese regions. Ultimately, we constructed JLSSD, a large-scale supervised dataset comprising 15 million 0.75-meter resolution images filtered from 10 million global grids (Figure 1). This includes 8.06 million quarterly image samples derived from 2.015 million Chinese locations and 7.985 million annual mosaic samples from 7.985 million non-Chinese locations. As illustrated in Figure 1, JLSSD demonstrates global coverage, diversity, temporal continuity, and spatial consistency, encompassing varied terrains and geomorphologies. To our knowledge, JLSSD represents the largest seasonal sub-meter-resolution self-supervised remote sensing dataset to date. Furthermore, JLSSD employs cluster-based data filtering to reduce redundant scene types (e.g., deserts, water bodies) and low-quality images, balancing inter-image diversity and intra-image heterogeneity. While increased heterogeneity challenges self-supervised image modeling, it ultimately enhances the feature representation capability of RSVFM.

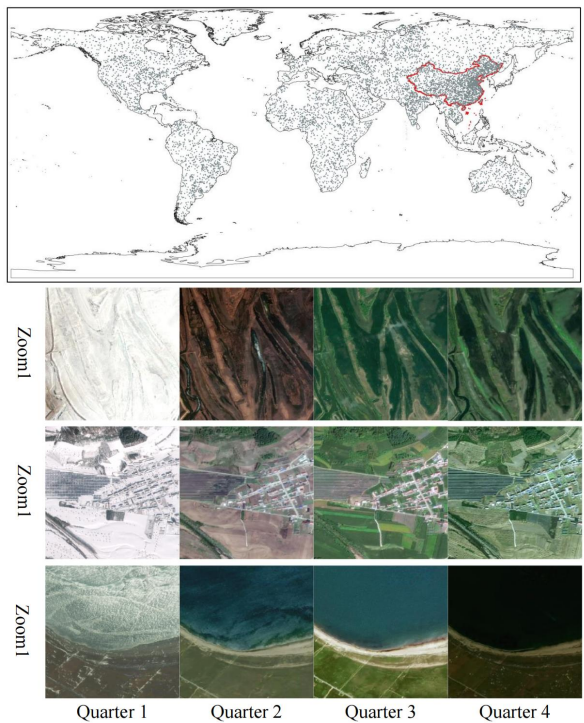

**Fig.1 JLSSD global distribution map**

## 2.2 Model Architecture

We propose a multi-granularity self-supervised learning framework for remote sensing imagery, as illustrated in Figure 2. The framework comprises three core modules: a data augmentation module for multi-view generation, a feature computation module for latent representation extraction, and a hybrid loss calculation module combining contrastive and reconstruction objectives. The pipeline operates as follows: An input image sample undergoes data transformations to generate 8 standard augmented views, 2 masked variants, and 3 seasonal contrastive views. The base image and its transformed variants are fed into a teacher-student framework, where the teacher and student models encode the images into latent representations. The framework jointly optimizes model parameters through cross-entropy loss for contrastive learning to align these latent representations.

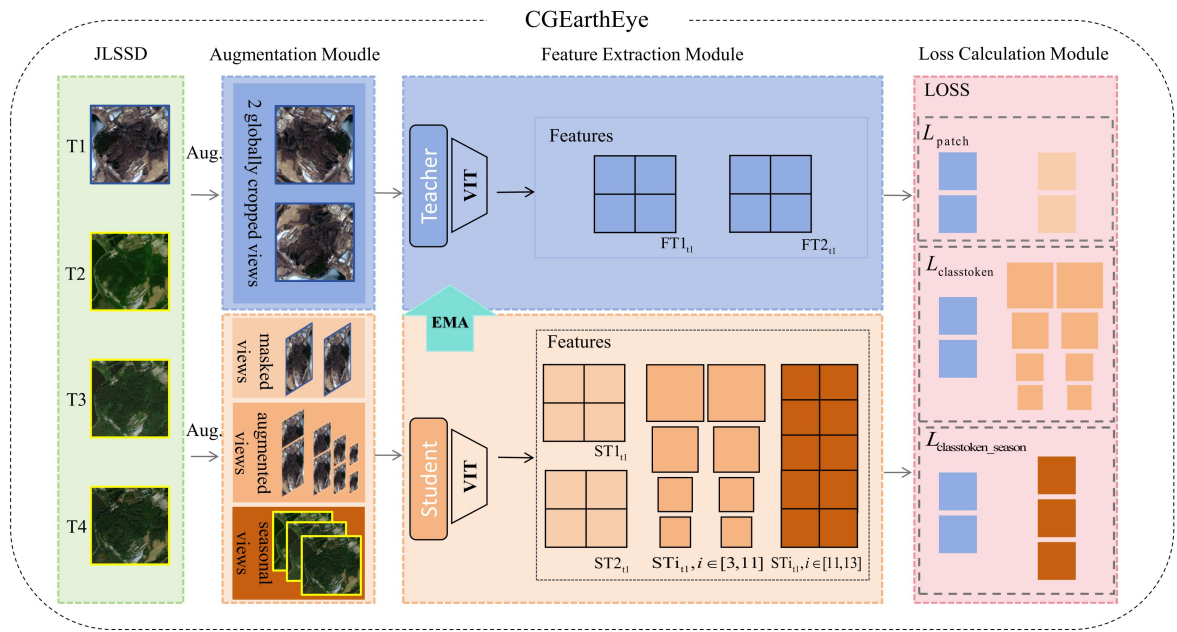

**Fig.2 Jilin-1 RSVFM pre-training framework diagram**

### 2.2.1 Augmentation Module

The essence of contrastive learning lies in aligning semantic consistency across different augmented views of the same image through model encoding. For an input image T1, the framework initiates a multi-scale cropping strategy that extracts both global and local regions at varying spatial scales, enhancing the model's capacity to integrate fine-grained local details with global contextual semantics. Following this, random color jittering—including adaptive adjustments to brightness, contrast, and saturation—is applied to improve robustness against illumination variations. Geometric transformations such as horizontal or vertical flipping are then introduced to diversify spatial representations. To support masked reconstruction tasks, block-wise masking (10% – 50% of pixel regions) is randomly applied to globally cropped images. Through this cascaded augmentation pipeline, T₁ generates 2 globally cropped views, 2 globally cropped views with optional masking, 8 multi-scale local crops, and 3 seasonal contrastive views for samples with quarterly temporal data. In total, each input image produces 15 augmented variants. The teacher model encodes the 2 global views to establish stable semantic anchors, while the student model processes the remaining 13 variants (local crops and seasonal views) to learn discriminative representations under diverse transformations. This asymmetric architecture ensures that semantic invariance is preserved through the teacher's guidance while encouraging the student to capture nuanced feature variations.

### 2.2.2 Feature Extraction Module

The feature extraction module primarily performs feature encoding on the output from the data augmentation module. It consists of both a teacher branch and a student branch, both employing the same ViT model [60], whose architecture is shown in Fig. 3. The Jilin-1 RSVFM has a total of 2.1 billion parameters and includes five ViT models of varying sizes, with parameter counts ranging from 22 million (22M) to 1.1 billion (1100M), to accommodate different application scenarios, as detailed in Table 1.

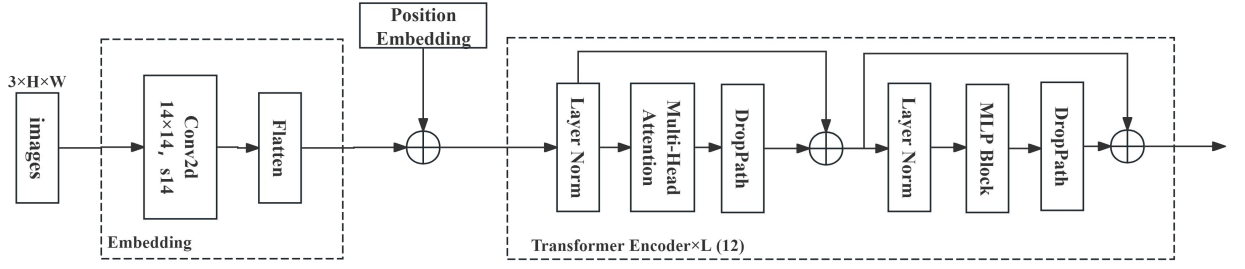

**Fig.3 VIT model architecture diagram**

**Table 1 CGEarthEye model parameter table**

| Model | Backbone | Layer Num | Embedding Dimension | Hidden Dimension | Attention Heads | Params. (M) |
|-------|----------|-----------|---------------------|------------------|-----------------|-------------|
| CGEarthEye-Small | VIT-S | 12 | 384 | 1536 | 6 | 22 |
| CGEarthEye-Base | VIT-B | 12 | 768 | 3072 | 12 | 86 |
| CGEarthEye-Large | VIT-L | 24 | 1024 | 4096 | 16 | 307 |
| CGEarthEye-Huge | VIT-H | 32 | 1280 | 5120 | 16 | 632 |
| CGEarthEye-Giant | VIT-G | 40 | 1536 | 6144 | 24 | 1100 |

The two globally cropped images are first fed into the teacher branch for encoding, yielding features $FT1_{t1}$、$FT2_{t1}$. The corresponding masked images are then input into the student branch for encoding, producing features $ST1_{t1}$、$ST2_{t1}$. Subsequently, three seasonal views and eight local crops are encoded using the student branch, resulting in feature sets $STi_{t1}, i \in [3,11]$、$STi_{t1}, i \in [11,13]$.

### 2.2.3 Loss Calculation Module

To model the model's global understanding capability of remote sensing imagery, we introduce cross-entropy loss for contrastive learning. The classification token features encoded by the two branches are transformed through a three-layer fully connected neural network, after which the loss is computed to perform augmentation-aware contrastive learning. The specific calculation is as follows.

$$L_{\text{classtoken}} = \sum_{T} \sum_{S} p_t \log p_s$$

Where $p_t$ denotes the class token from the fully connected layer for the two globally augmented images processed by the teacher branch. $p_s$ denotes the class token from the fully connected layer for the eight locally augmented images processed by the student branch.

To model the representation capability of the model for multi-seasonal imagery, we further apply the contrastive loss to three locally cropped patches over Chinese regions. The specific calculation is detailed below.

$$L_{\text{classtoken\_season}} = \sum_T \sum_S p_t \log p_{s\_season}$$

Where $p_{s\_season}$ denotes the class token from the fully connected layer for the three seasonal views processed by the student branch.

To model the pixel-level prediction capability of the model for remote sensing imagery, we extend the contrastive loss calculation to encoded features of masked patches. These features are supervised using outputs from corresponding locations in the teacher branch, with detailed computations specified below.

$$L_{\text{patch}} = \sum_i p_{ti} \log p_{si}$$

Where $i$ denotes index of the masked patches. The outputs from the corresponding positions of the teacher network are used to supervise the student network, as shown below.

The final loss function is computed as follows.

$$L = L_{\text{classtoken}} + L_{\text{classtoken\_season}} + L_{\text{patch}}$$

After computing the loss through forward propagation, the model performs backpropagation to calculate gradients. During backpropagation, only the parameters of the student branch are activated. These parameters are then updated using the Stochastic Gradient Descent (SGD) algorithm. For the teacher branch parameters, an Exponential Moving Average (EMA) momentum update strategy [61] is applied to prevent model collapse during training. The detailed computation is specified below.

$$\theta_t = m \times \theta_{t-1} + (1-m) \times \theta_s$$

where $\theta_t$、$\theta_s$ denote the parameters of the teacher branch and student branch at the current time step, respectively. $\theta_{t-1}$ denote the parameters of the teacher branch at the last time step, and $m$ is a momentum coefficient with 0.992 in this study.

## 3 Experiments and analysis

### 3.1 Pre-training Implementation

To accelerate model training, we incorporate multiple optimization techniques across the full training pipeline. At the data level, LMDB is utilized for storage and management of training data to enhance loading efficiency. For the model architecture, the FlashAttention algorithm [62] accelerates attention computation. Regarding training strategy, Fully Sharded Data Parallelism (FSDP) [63] shards model, optimizer, and gradient parameters while enabling mixed precision training, effectively increasing batch size. Inspired by DINOv2 [36], we first train the model using $224 \times 224$ global crops and $98 \times 98$ local crops, then scale global crops to $518 \times 518$. Ultimately, pretraining efficiency improved by approximately $2 \times$ with 60% GPU memory reduction compared to the baseline.

The experiment was conducted over 150 days using 16 NVIDIA A800 GPUs (80GB), with the hardware environment detailed in Table 2.

**Table 2 Experiment hardware and software environment configuration**

| Experimental Environment | Configuration |
|---|---|
| | CPU: 2× Intel 8358P 2.6GHz / 32-core / 48MB / 240W |
| | GPU: 16× NVIDIA A800 GPU / 80GB VRAM |
| Hardware Environment | RAM: 2TB DDR4 (32 slots × 32GB × 2 channels) |
| | Storage: 30.72TB NVMe SSD (4× 7.68TB) |
| | Network: 4× 200G InfiniBand + 1× 100G InfiniBand |
| Soft Environment | Ubuntu 20.04.5 LTS |
| Coding Environment | Visual Studio Code |
| Framework | Pytorch2.0.0 |

## 3.2 Performance in downstream task

This study comprehensively evaluates CGEarthEye's performance on four classic remote sensing tasks—scene classification, object recognition, semantic segmentation, and change detection—using frozen backbone fine-tuning. We employ the most representative and widely used benchmark datasets from the literature, comparing results with other remote sensing foundation models.

### 3.2.1 Scene Classification

We first assess the pretrained model on scene classification tasks, which requires no extra decoder and directly reflects the model's overall representation capability.

1) Dataset

RESISC-45 [64]：A scene classification dataset from Northwestern Polytechnical University. It contains 31,500 images across 45 classes (700 samples/class), with uniform 256×256 pixel resolution.

AID [22]：A benchmark dataset from Wuhan University for high-resolution remote sensing interpretation. It includes 10,000 images spanning 30 land-cover categories at 600 × 600 pixels, with spatial resolutions ranging from 0.5 to 8 meters.

2) Implementation Details

All experiments for scene classification are conducted within the MMPretrain framework, with identical hyperparameters applied to both RESISC-45 and AID datasets. The training configuration uses a batch size of 64 over 200 epochs, an initial learning rate of 1e-6, and the AdamW optimizer with cosine annealing scheduling. Data augmentation employs RandomResizedCrop and RandomFlip, while input images are uniformly resized to 224× 224 pixels. A linear classifier serves as the classification head, with parallel training under both frozen and activated backbone settings.

3) Finetuning Results

As shown in Table 4, CGEarthEye significantly outperforms existing remote sensing foundation models (e.g., SkySense) on both datasets, achieving state-of-the-art (SOTA) accuracy. Notably, even with a frozen backbone (only optimizing the linear classifier), CGEarthEye consistently surpasses other vision foundation models.

Table 3 CGEarthEye experimental results of scene classification（* indicates training with frozen backbone）

| Method | Backbone | RESISC-45 | AID |
|---|---|---|---|
| | | OA | OA |
| SeCo[48] | ResNet50 | 0.9291 | 0.9347 |
| GASSL[46] | ResNet50 | 0.9306 | 0.9355 |
| CACo[47] | ResNet50 | 0.9194 | 0.9088 |
| SatLas*[65] | Swin-B | - | 0.6598 |
| SatLas | Swin-B | 0.9470 | 0.9496 |
| CMID*[54] | Swin-B | - | 0.8780 |
| CMID | Swin-B | 0.9553 | 0.9611 |
| RingMo[50] | Swin-B | 0.9567 | 0.9690 |
| GFM*[33] | Swin-B | - | 0.7942 |
| GFM | Swin-B | 0.9464 | 0.9547 |
| SatMAE[51] | VIT-L | 0.9410 | 0.9502 |
| Scale-MAE*[52] | ViT-L | - | 0.7643 |
| Scale-MAE | ViT-L | 0.9504 | 0.9644 |
| SSL4EO[24] | ViT-B | 0.9127 | 0.9106 |
| RVSA[18] | ViT-B | 0.9569 | 0.9703 |
| SkySense*[57] | Swin-H | - | 0.9407 |
| SkySense | Swin-H | 0.9632 | 0.9768 |
| MTP[66] | InternImage-XL | 0.9627 | - |
| CGEarthEye* | VIT-G | 0.9584 | 0.9760 |
| CGEarthEye | VIT-G | 0.9675 | 0.9769 |

### 3.2.2 Object Detection

Following scene-level recognition tasks, this section focuses on object-level detection, evaluating both horizontal and rotated bounding box detection performance on the DIOR [67] and DIOR-R [68] datasets.

1) Dataset

DIOR is a benchmark dataset for multi-scale object detection in complex scenarios, jointly released by Wuhan University and the Aerospace Information Research Institute. It contains 23,463 images with $800 \times 800$ pixels across 20 object categories with 192,472 annotated instances, featuring spatial resolutions from 0.5 to 30 meters.

DIOR-R is an extended version designed for rotated object detection, facilitating precise localization of arbitrarily oriented targets in remote sensing imagery.

2) Implementation Details

For horizontal box detection (DIOR), we fine-tune models using the DINO detector head (H. Zhang et al.,

2022) within the MMDetection framework (Chen et al., 2019). The training configuration uses a batch size of 4 over 60 epochs, an initial learning rate of 1e-4 and the AdamW optimizer with cosine annealing scheduling. Data augmentation employs RandomResizedCrop and RandomFlip, while input images are uniformly resized to $784 \times 784$ pixels. For rotated box detection (DIOR-R), the RHINO head [68] in MMRotate is adopted with identical hyperparameters except for a reduced batch size of 2.

3) Finetuning Results

CGEarthEye achieves superior results on both tasks, attaining mAPs of 0.8262 (DIOR) and 0.7520 (DIOR-R), surpassing all compared remote sensing foundation models, including SkySense and MTP. Crucially, these state-of-the-art results are achieved via frozen backbone fine-tuning, demonstrating that CGEarthEye's pretrained backbone captures transferable object-level representations enabling high-precision detection with minimal adaptation.

**Table 4 CGEarthEye experimental results of object detection (* indicates training with frozen backbone)**

| Method | Backbone | DIOR | DIOR-R |
| --- | --- | --- | --- |
| | | mAP | mAP |
| GASSL[46] | ResNet50 | 0.6740 | 0.6565 |
| CACo[48] | ResNet50 | 0.6691 | 0.6410 |
| SatLas[65] | Swin-B | 0.7410 | 0.6759 |
| CMID[54] | Swin-B | 0.7511 | 0.6637 |
| RingMo[50] | Swin-B | 0.7590 | -- |
| GFM[33] | Swin-B | 0.7284 | 0.6767 |
| SatMAE[51] | VIT-L | -- | 0.6566 |
| Scale-MAE[52] | ViT-L | 0.7381 | 0.6647 |
| SSL4EO[24] | ViT-B | 0.6482 | 0.6123 |
| RVSA[18] | ViT-B | 0.7322 | 0.7105 |
| SkySense[57] | Swin-H | 0.7873 | 0.7427 |
| MTP[66] | ViT-L+RVSA | 0.8110 | 0.7454 |
| CGEarthEye* | VIT-G | 0.8262 | 0.7520 |

### 3.2.3 Semantic Segmentation

To evaluate CGEarthEye's fine-tuning performance on finer-grained pixel-level tasks, this section assesses its semantic segmentation capability. Semantic segmentation is a critical application for land cover and object recognition in remote sensing.

1) Dataset

LOVEDA is an open-source benchmark for land cover classification and cross-domain adaptation (Wuhan University). It comprises 5,982 high-resolution patches at $1024 \times 1024$ pixels and 0.3m resolution with 7 semantic classes [69].

iSAID is an aerial imagery instance segmentation benchmark (Wuhan University & ISPRS). It containsg 2,806 images at $800 \times 800$ to $13,000 \times 11,000$ pixels with 0.3 – 1.5m resolution, and labels 655,451 instances across 15 categories [70].

Potsdam is an open benchmark dataset released by the International Society for Photogrammetry and Remote Sensing (ISPRS) used for semantic segmentation research of high-resolution remote sensing imagery. It comprises 38 orthorectified aerial images with 6 categories of fine-grained semantic labels. Each image measures 6000 × 6000 pixels and offers a spatial resolution as high as 0.05 meters [49].

2)Implementation Details

All experiments for semantic segmentation are conducted within the MMSegmentation framework. Data processing follows SkySense[57] and MTP[66]. The training configurations of LoveDA, iSAID and Potsdam are consistent. During training, we use a batchsize of 8, an initial learning rate of 1e-6 and the AdamW optimizer with cosine annealing scheduling. Data augmentation employs RandomResizedCrop and RandomFlip, while input images are uniformly resized to 518 × 518pixels. The UperNet segmentation head is deployed with frozen backbone training due to computational constraints.

3)Finetuning Results

The fine-tuning results for semantic segmentation are presented in Table 5. Experimental findings demonstrate that CGEarthEye effectively enhances the performance of foundational remote sensing models on semantic segmentation tasks. On the LoveDA dataset, CGEarthEye achieves state-of-the-art (SOTA) performance with a mean Intersection over Union (mIoU) of 56.67%, surpassing the 54.17% obtained by the fully fine-tuned MTP model. On the iSAID and Potsdam datasets, its mIoU is 1.4% and 0.46% lower than the fully fine-tuned SkySense model, respectively. It is noteworthy that SkySense was trained on over 21.5 million pairs of multimodal data, utilizing geolocation awareness, multimodal and multitemporal contrastive learning, and leveraging more than 80 servers each equipped with 8 A100 GPUs. Consequently, its overall training cost significantly exceeds that of CGEarthEye. Notably, it exceeds the frozen accuracy of SkySense by 4.11% and outperforms all other foundational vision models except for the fully fine-tuned SkySense.

**Table 5 CGEarthEye experimental results of semantic segmentation (* indicates training with frozen backbone)**

| Method | Backbone | LoveDA | iSAID | Potsdam |
|---|---|---|---|---|
| | | mIoU | | mF1 |
| SeCo[48] | ResNet50 | 0.4363 | 0.5720 | 0.8903 |
| GASSL[46] | ResNet50 | 0.4876 | 0.6595 | 0.9127 |
| CACo[47] | ResNet50 | 0.4889 | 0.6432 | 0.9135 |
| SatLas*[65] | Swin-B | - | 0.5603 | - |
| SatLas | Swin-B | - | 0.6871 | 0.9128 |
| CMID*[54] | Swin-B | - | 0.5940 | - |
| CMID | Swin-B | - | 0.6621 | 0.9186 |
| RingMo[50] | Swin-B | - | 0.6720 | 0.9127 |
| GFM*[33] | Swin-B | - | 0.6086 | - |
| GFM | Swin-B | - | 0.6662 | 0.9185 |
| Scale-MAE*[51] | ViT-L | - | 0.6577 | - |
| Scale-MAE | ViT-L | - | 0.4653 | 0.9154 |
| SSL4EO[24] | ViT-B | - | 0.6401 | 0.9154 |
| RVSA[18] | ViT-B | 0.5244 | 0.6449 | - |
| SkySense*[57] | Swin-H | - | 0.6540 | - |
| SkySense | Swin-H | - | 0.7091 | 0.9399 |
| MTP[66] | InternImage-XL | 0.5417 | - | |
| CGEarthEye* | VIT-G | 0.5667 | 0.6951 | 0.9353 |

### 3.2.4 Change Detection

Finally, we focus on the change detection task, which identifies temporal change features in co-registered remote sensing (RS) imagery by modeling it as a specialized segmentation problem. This section specifically examines the most representative bitemporal change detection paradigm.

1) Dataset

LEVIR-CD is an open benchmark dataset for building-scale land change detection, comprising 637 bitemporal RS image pairs with temporal spans of 5-14 years. It employs binary semantic annotations, with image dimensions of $1024 \times 1024$ pixels and a spatial resolution of 0.5m[71].

SYSU-CD is an open-source benchmark dataset for multi-category change detection in complex urban scenes, containing 12,000 bitemporal RS image pairs. It features 5-class semantic change annotations, temporal spans of 3-8 years, image dimensions of $512 \times 512$ pixels, and a spatial resolution of 0.8m[72].

CDD is a versatile open-source dataset for multi-scale land cover change detection, consisting of 16,000 bitemporal RS image pairs. It adopts a multi-level annotation scheme including binary change masks, 6-class semantic change labels, and change driver tags. Image dimensions range from 256 to 4096 pixels with spatial resolutions between 0.1m and 2m. This study specifically tests on binary change annotations [73]。

2) Implementation Details

The change detection experiments are implemented using the Open-CD framework. The model architecture adopts Changeformer [74], with hyperparameter settings consistent with those used in the semantic segmentation tasks.

3) Finetuning Results

As shown in Table 6, the fine-tuning results demonstrate that CGEarthEye effectively enhances the performance of foundational remote sensing models on change detection tasks. With frozen backbone fine-tuning, CGEarthEye achieves optimal or suboptimal accuracy across all three change detection datasets: it ranks first on SYSU-CD, outperforming other fully fine-tuned models; places third on LEVIR-CD with its performance 0.21% and 0.12% lower than MTP and SkySense respectively; and trails the fully fine-tuned MTP by 0.33% on CDD. Notably, given the sophisticated pretraining configurations of MTP and SkySense, CGEarthEye's frozen backbone fine-tuning attains comparable change detection accuracy to these models, demonstrating superior generalization capability and robustness.

**Table 6 CGEarthEye experimental results of change detection (* indicates training with frozen backbone)**

| Method | Backbone | LEVIR-CD | SYSU-CD | CDD |
|---|---|---|---|---|
| | | F1 | F1 | F1 |
| ChangeFormer[74] | MiT-B2 | 0.9111 | 0.8311 | - |
| BiT-18[75] | ResNet-18 | 0.8931 | - | - |
| STANet[71] | - | - | 0.7736 | - |
| HANet[76] | ResNet-101 | 0.9028 | 0.7741 | 0.8923 |
| CGNet[77] | VGG-16 | 0.9201 | 0.7992 | 0.9473 |
| SGSLN[78] | - | 0.9233 | 0.8307 | 0.9624 |
| C2FNet[79] | VGG-16 | 0.9183 | 0.7797 | 0.9593 |
| MutSimNet[80] | - | 0.9200 | 0.8234 | |
| CACG-Net[81] | - | 0.9229 | 0.8335 | 0.9473 |
| MTP[66] | InternImage-XL | 0.9267 | - | 0.9837 |
| SkySense[57] | Swin-H | 0.9258 | - | - |
| ChangeClip[19] | ViT-B | 0.9201 | 0.8332 | 0.9789 |
| CGEarthEye* | VIT-G | 0.9246 | 0.8347 | 0.9804 |

# 4 Discussion

This section comprehensively investigates and discusses the characteristics of CGEarthEye, with emphasis on feature extraction efficacy, parameter volume, fine-tuning strategies, comparison with visual foundation models, and spatial distribution mapping performance.

## 4.1 Pretrained Feature Visualization

To evaluate the feature representation capability of CGEarthEye, we employ Principal Component Analysis (PCA) transformation and K-means clustering for feature visualization. Specifically, a $518 \times 518 \times 3$ input image processed through the ViT-G model yields a $37 \times 37 \times 6144$ feature map. The top three principal components by contribution rate are visualized in true color via PCA, while the top ten principal components undergo K-means clustering to generate 3-5 categorical outputs, as illustrated in Figure 4. Across ten distinct terrain scenarios, PCA-derived features consistently delineate primary object boundaries within the imagery. Concurrently, K-means clustering effectively extracts foreground features including buildings, factories, roads, croplands, and water bodies. Collectively, these results demonstrate CGEarthEye's superior feature representation capability, providing robust support for downstream applications.

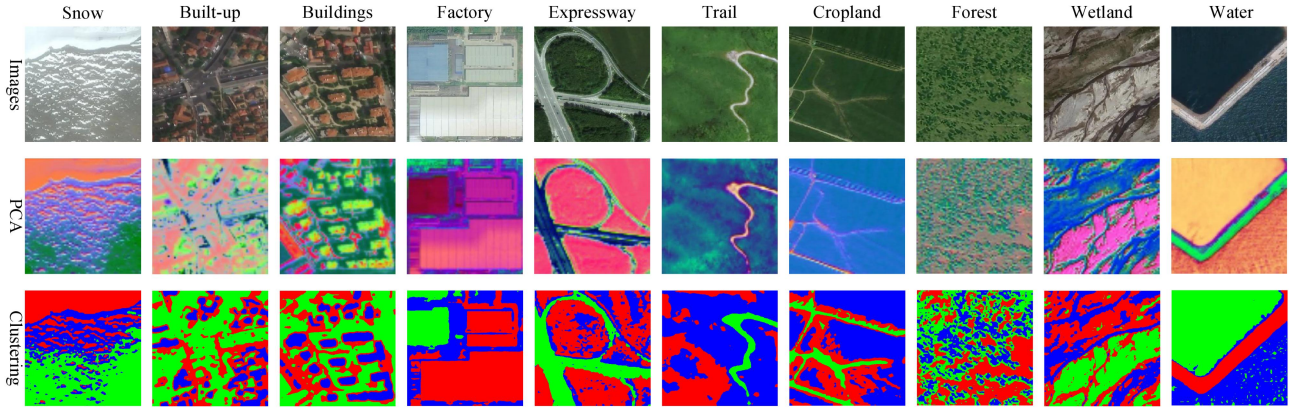

**Fig.4 Visualization analysis of PCA and clustering for CGEarthEye model features**

## 4.2 Impact of Parameter Scale on Model Performance

ViTs establish a task-agnostic universal representation paradigm through unified global self-attention mechanisms and hierarchical feature encoding architectures, enabling robust cross-task and cross-dataset generalization. In this study, we construct five remote sensing foundation models with varying parameter scales under the CGEarthEye framework, using ViT-S, ViT-B, ViT-L, ViT-H, and ViT-G as backbones. To investigate parameter scaling effects, we conduct detailed evaluations on remote sensing image scene classification tasks (Table 7). Results indicate progressive performance improvement on three benchmark datasets, RESISC-45 [64], AID [22], and fMoW [25], as model parameters increase. Notably, this scaling behavior exhibits dataset-dependent patterns correlated with task difficulty. On the most challenging fMoW dataset, accuracy rises from 0.8421 for CGEarthEye-Small (22M parameters) to 0.9298 for CGEarthEye-Giant (1100M parameters), constituting an 8% absolute gain. Conversely, parameter saturation emerges on less complex datasets, where marginal differences are observed between CGEarthEye-Large (307M), CGEarthEye-Huge (632M), and CGEarthEye-Giant (1100M) models on RESISC-45 and AID. These findings confirm that scaling model size effectively enhances feature extraction capacity for rapid performance gains, with improvement magnitude positively correlated with task complexity [57]. However, the observed accuracy saturation indicates that ultra-large parameter models are not universally required. CGEarthEye's scalable parameter configuration facilitates adaptable deployment across diverse downstream applications.

**Table 7 Performance of CGEarthEye models with varying parameter scales in image classification tasks (\* indicates training with frozen backbone)**

| Model | Backbone | RESISC-45 | AID | fMoW |
| --- | --- | --- | --- | --- |
| | | OA | OA | OA |
| CGEarthEye-S* | ViT-S | 0.9070 | 0.9438 | 0.4861 |
| CGEarthEye-S | ViT-S | 0.9608 | 0.9620 | 0.8421 |
| CGEarthEye-B* | ViT-B | 0.9308 | 0.9581 | 0.5612 |
| CGEarthEye-B | ViT-B | 0.9668 | 0.9759 | 0.8853 |
| CGEarthEye-L* | ViT-L | 0.9542 | 0.9761 | 0.7756 |
| CGEarthEye-L | ViT-L | 0.9676 | 0.9763 | 0.8871 |
| CGEarthEye-H* | ViT-H | 0.9563 | 0.9762 | 0.7815 |
| CGEarthEye-H | ViT-H | 0.9674 | 0.9766 | 0.9012 |
| CGEarthEye-G* | ViT-G | 0.9584 | 0.9760 | 0.8980 |
| CGEarthEye-G | ViT-G | 0.9675 | 0.9769 | 0.9298 |

## 4.3 Impact of Frozen Backbone on Model Performance

Given computational constraints, our primary evaluations across four downstream tasks were conducted with frozen backbones. As demonstrated in Section 3.2, CGEarthEye's superior feature extraction capability enables performance comparable to fully fine-tuned state-of-the-art models like SkySense and MTP under frozen-backbone settings. We quantitatively analyze performance differences between frozen and fully fine-tuned configurations on image classification tasks (Table 7). Across all three classification datasets, every CGEarthEye variant exhibits higher accuracy in fully fine-tuned mode than in frozen-backbone mode. This indicates that unfreezing parameters during fine-tuning can further unlock CGEarthEye's potential. Notably, the performance degradation from backbone freezing varies significantly across datasets and model scales. On the challenging fMoW dataset, frozen-backbone CGEarthEye-Giant shows >2% accuracy drop versus full fine-tuning, while the gap is narrower (<0.8%) on other datasets. More dramatically, CGEarthEye-Small suffers over 35% accuracy degradation when frozen on fMoW, with varying but substantial gaps on other datasets. ntegrating findings from Section 3.2 and Table 7 reveals that CGEarthEye achieves exceptional accuracy with frozen-backbone fine-tuning, enabling high-performance downstream adaptation at low computational cost. When sufficient GPU resources are available, full parameter fine-tuning delivers additional performance gains.

## 4.4 Convergence Efficiency Comparison

Convergence speed on downstream tasks is a critical metric for evaluating foundation models. Fundamentally, effective pretraining that learns robust feature representations accelerates convergence and enhances overall task performance. We benchmark convergence efficiency against DINOv2 and smaller models on three scene classification datasets. Training convergence curves are presented in Figure 5. Results show CGEarthEye achieves superior convergence rates across all datasets under identical experimental setups. This accelerated convergence demonstrates that our pretraining effectively captures and encodes discriminative feature representations, enabling rapid adaptation to downstream tasks.

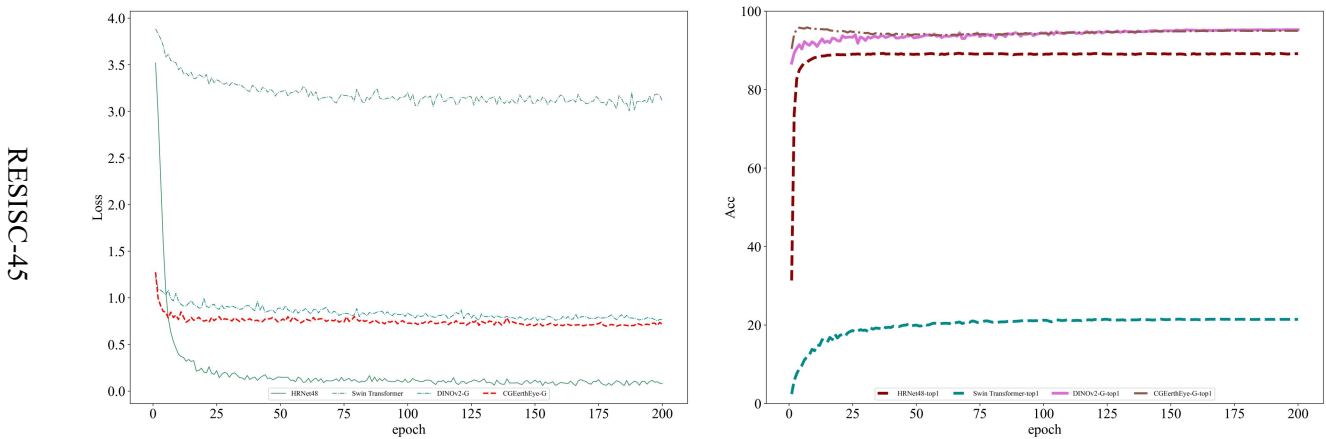

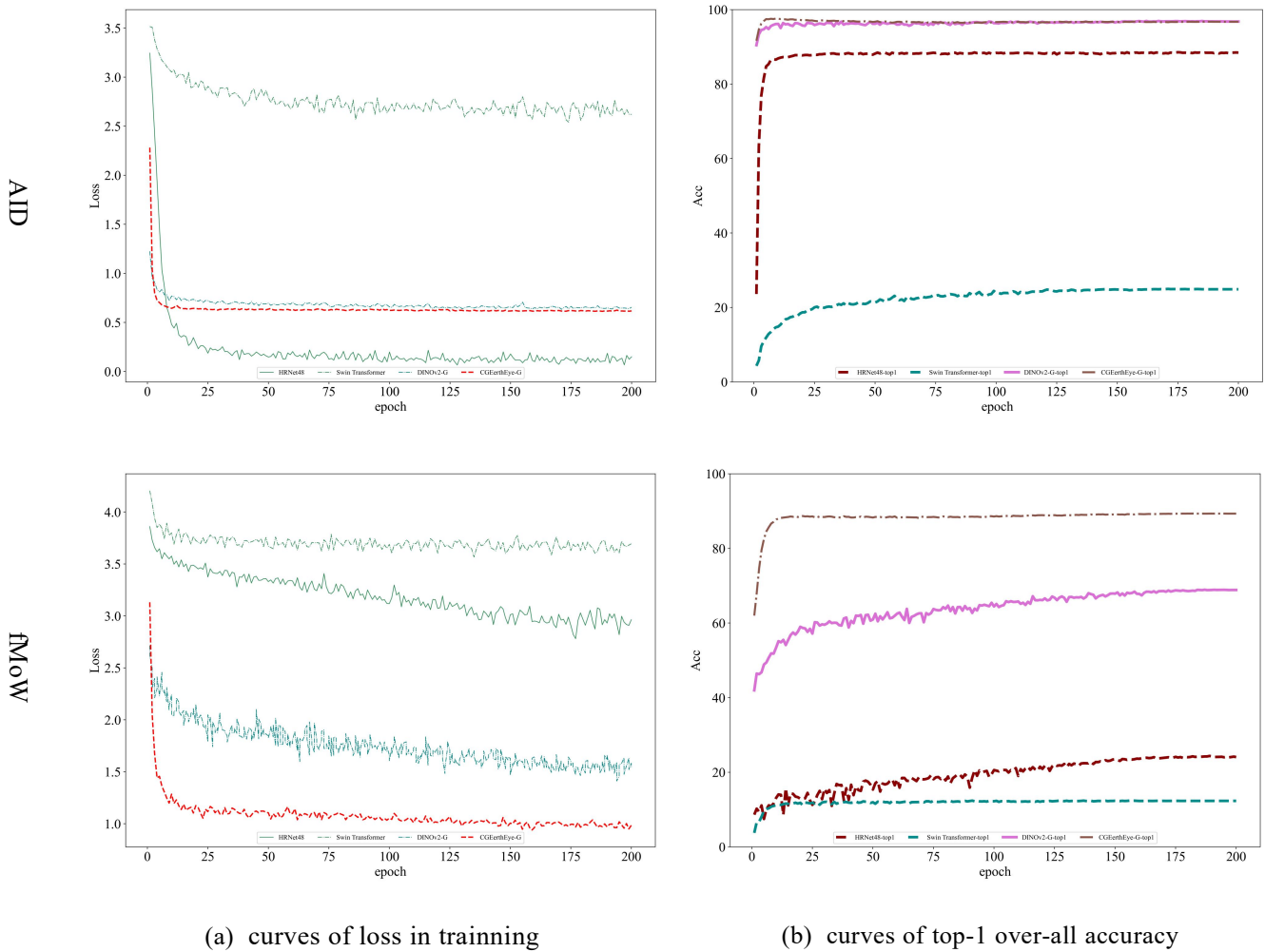

<div align="center">

(a) curves of loss in trainning        (b) curves of top-1 over-all accuracy

</div>

**Fig.5 Convergence curves of different methods on RESISC-45, AID and fMoW datasets: (a) curves of loss in trainning, (b)curves of top-1 over-all accuracy in testing dataset**

4.5 Comparison Between Remote Sensing and Natural Image Foundation Models

Section 3.2 demonstrates the significant advantage of CGEarthEye pretrained models over randomly initialized models trained from scratch. Furthermore, we systematically compare CGEarthEye with the state-of-the-art computer vision foundation model DINOv2 across diverse Earth observation tasks. Specifically, we evaluate frozen-backbone fine-tuning on one representative dataset per task category (image classification, object detection, semantic segmentation, and change detection), with results detailed in Table 8. CGEarthEye consistently outperforms DINOv2 by significant margins across all four tasks. This performance discrepancy may be attributed to two key factors. The substantial domain gap between remote sensing imagery and natural images hinders effective transfer learning for models like DINOv2 pretrained exclusively on natural images. DINOv2 lacks specialized architectural designs for remote sensing characteristics, particularly in capturing spatiotemporal features inherent to RS data. Consequently, DINOv2 fails to leverage the rich spatiotemporal attributes of RS imagery for downstream tasks. In contrast, CGEarthEye — explicitly designed for remote sensing — integrates large-scale RS-specific pretraining data, methodologies, and model architectures that inherently align with downstream interpretation tasks. This domain-specific optimization yields significantly superior performance.

**Table 8 Performance comparison between CGEarthEye and DINOv2 in downstream tasks (\* indicates training with frozen backbone)**

| Model | Backbone | RESISC-45 | DIOR | LoveDA | SYSU-CD |
| --- | --- | --- | --- | --- | --- |
| | | OA | mIoU | mIoU | F1 |
| DINOv2-G*[36] | ViT-G | 0.9529 | 0.8020 | 0.5514 | 0.8159 |
| CGEarthEye-G* | ViT-G | 0.9584 | 0.8262 | 0.5667 | 0.8347 |

4.6 Applications for Spatial Distribution Mapping of Geographic Features

Spatial distribution mapping of geographic features represents a primary application of remote sensing imagery, where performance directly determines the utility level of vision foundation models in downstream implementations. To comprehensively evaluate CGEarthEye's regional-scale spatial mapping capability, we conduct case studies in Longhua District, Shenzhen, focusing on three practical tasks: building extraction, crane detection, and comprehensive change detection.

1)  Crane Detection

The crane detection model combines CGEarthEye with a DINO detection head, trained exclusively on the Jilin-1 Crane Detection Dataset. For comparative analysis, we established a YOLOv8 baseline model. The Jilin-1 dataset contains 24,887 image-label pairs featuring sub-meter resolution imagery captured across major Chinese cities. Both models processed 0.75-meter resolution satellite imagery of Longhua District from the third quarter of 2023. Accuracy assessment employed quadrat sampling methodology. As Table 9 demonstrates, CGEarthEye identified 221 cranes compared to YOLOv8's 234 detections. CGEarthEye surpassed YOLOv8 universally across evaluation metrics: achieving 0.9457 precision exceeding YOLOv8 by 6.21%, 0.8261 recall surpassing YOLOv8 by 7.89%, and a 0.8818 F1-score outperforming YOLOv8 by 7.2%. These results highlight CGEarthEye's exceptional proficiency in extracting small objects such as construction cranes, attributable to its advanced representation learning framework optimized for fine-grained object recognition.

**Table 9 Comparison of tower crane detection in Longhua District, Shenzhen.**

| Model | Detections | Ground Truths | Recall | Precision | F1 |
| --- | --- | --- | --- | --- | --- |
| YOLOv8[82] | 213 | 253 | 0.7472 | 0.8836 | 0.8096 |
| CGEarthEye-B | 221 | 253 | 0.8261 | 0.9457 | 0.8818 |

Visual comparisons of crane detection results in Longhua District are presented in Figure 6. While both models exhibit consistent spatial distribution patterns, CGEarthEye demonstrates superior recall by identifying cranes missed by YOLOv8. Close-up views reveal comparable performance for cranes with high target-background contrast. However, in complex scenarios with low chromatic contrast (e.g., cranes against bare backgrounds), YOLOv8 exhibits significant omission errors whereas CGEarthEye maintains robust detection capability.

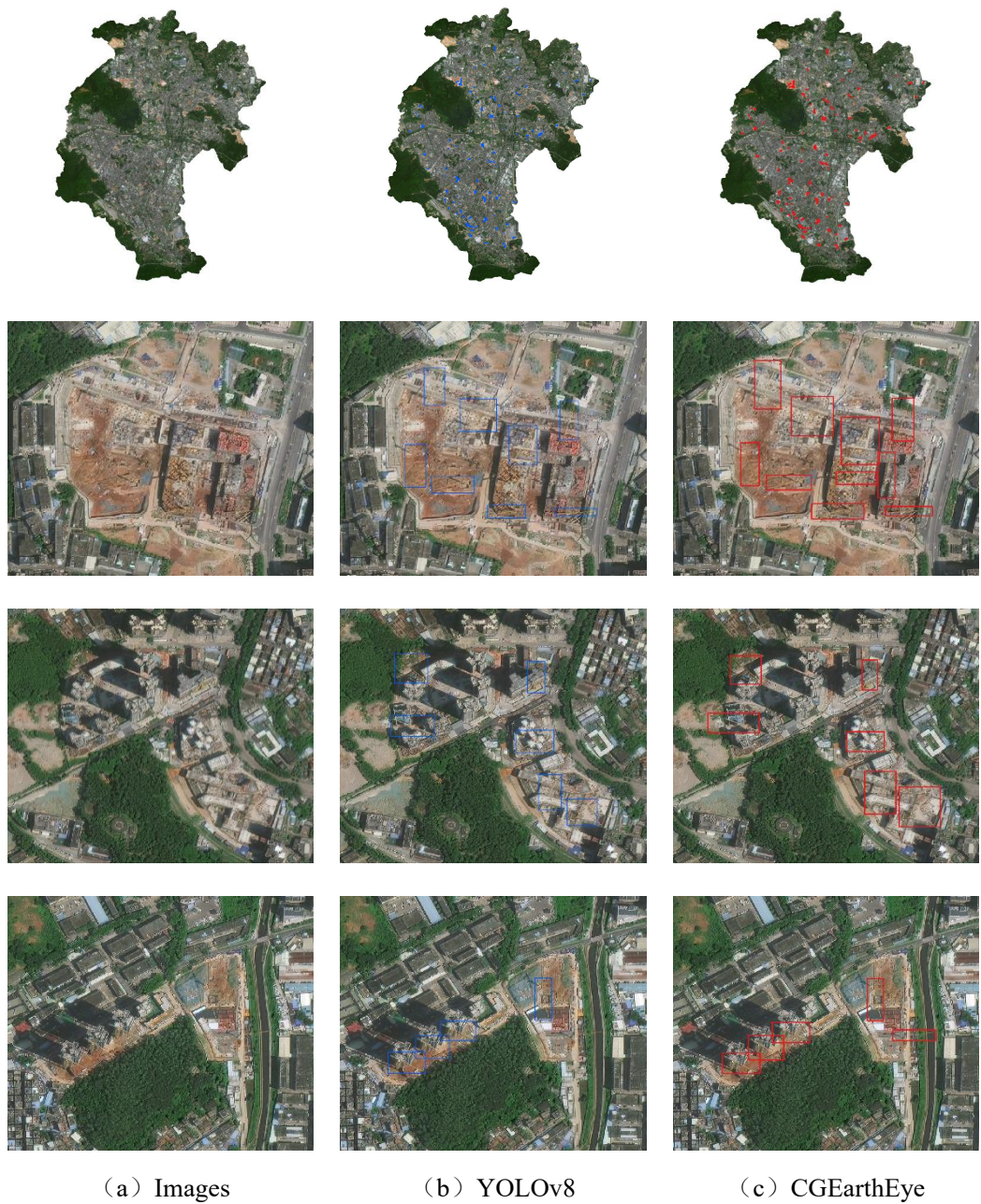

（a）Images　　　　　　　（b）YOLOv8　　　　　　　（c）CGEarthEye

**Fig.6 Comparative visualization of tower crane detection in Longhua District, Shenzhen**

2) Building Extraction

Building extraction holds significant value for urban planning, disaster management, and land monitoring. As a canonical semantic segmentation task, we implement CGEarthEye with UperNet framework trained on the Jilin-1 Building Extraction Dataset, using Swin Transformer [83] as benchmark. The dataset contains 27,000 sub-meter resolution image-mask pairs covering urban agglomerations and rural settlements across major Chinese cities. Applied to 0.75m resolution imagery of Longhua District (Q3 2023), model performance was evaluated via quadrat sampling. Results (Table 10) show CGEarthEye detected 37,761 building footprints, with 7,410 more than Swin Transformer's 30,351. CGEarthEye outperformed Swin Transformer across all metrics: achieving 19% higher recall, marginally superior precision (>0.92 vs 0.91), and 12.3% higher F1-score (0.873 vs 0.777). This demonstrates CGEarthEye's exceptional adaptability to regional-scale building extraction despite its compact architecture.

**Table 10 Comparison of building extraction in Longhua District, Shenzhen**

| Model | Detections | Ground Truths | Recall | Precision | F1 |
|---|---|---|---|---|---|
| Swin Transformer | 30351 | 38857 | 0.7725 | 0.9890 | 0.8332 |
| CGEarthEye-B | 37761 | 38857 | 0.9654 | 0.9934 | 0.9279 |

Visual comparisons of building extraction results in Longhua District are presented in Figure 7. Despite comparable patch-level accuracy between models, CGEarthEye demonstrates markedly superior pixel-wise classification capability. Compared to Swin Transformer, CGEarthEye achieves more precise boundary delineation and contour extraction for buildings, effectively classifying ambiguous edge regions with minimal over-segmentation or under-segmentation artifacts. Furthermore, CGEarthEye substantially outperforms Swin Transformer in complex scenarios involving vacant lots, basketball courts, and low-rise industrial buildings.

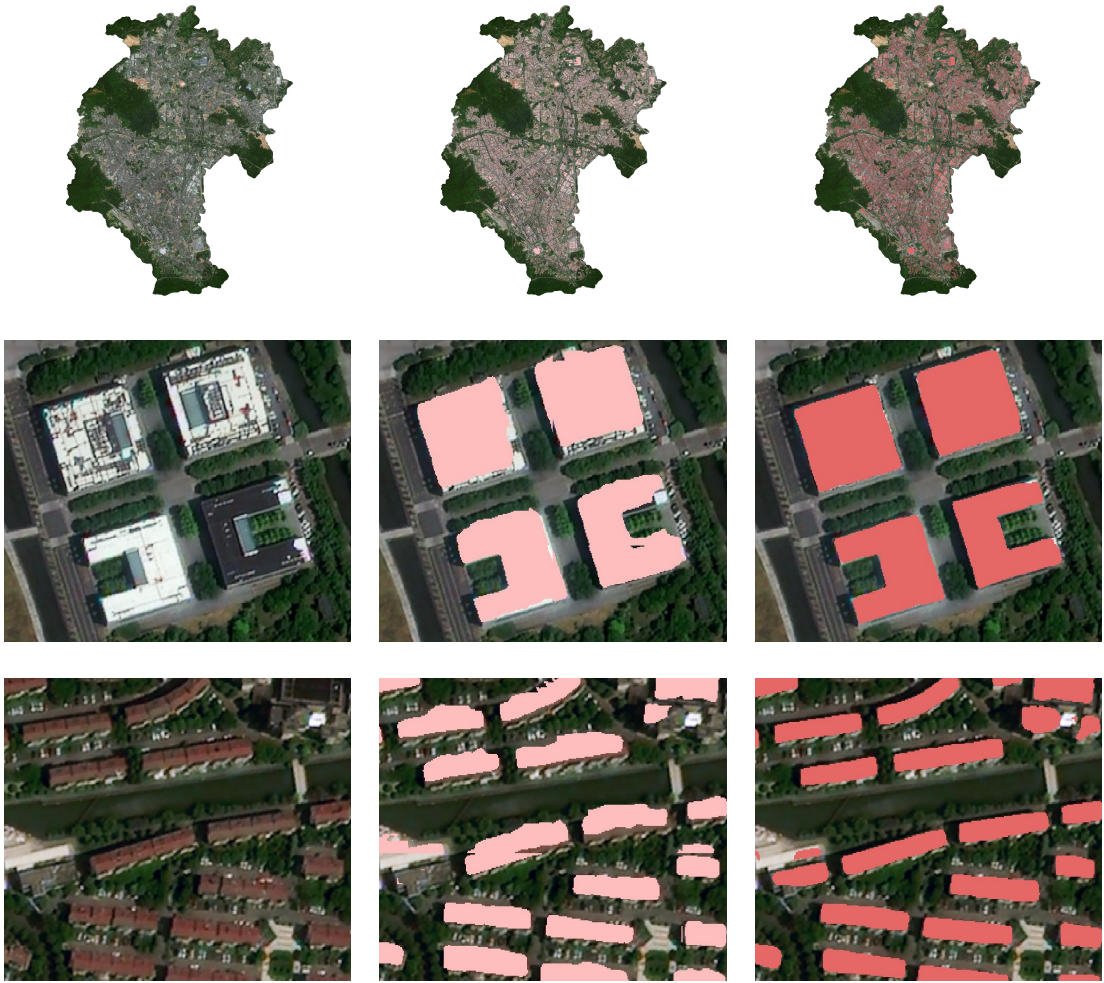

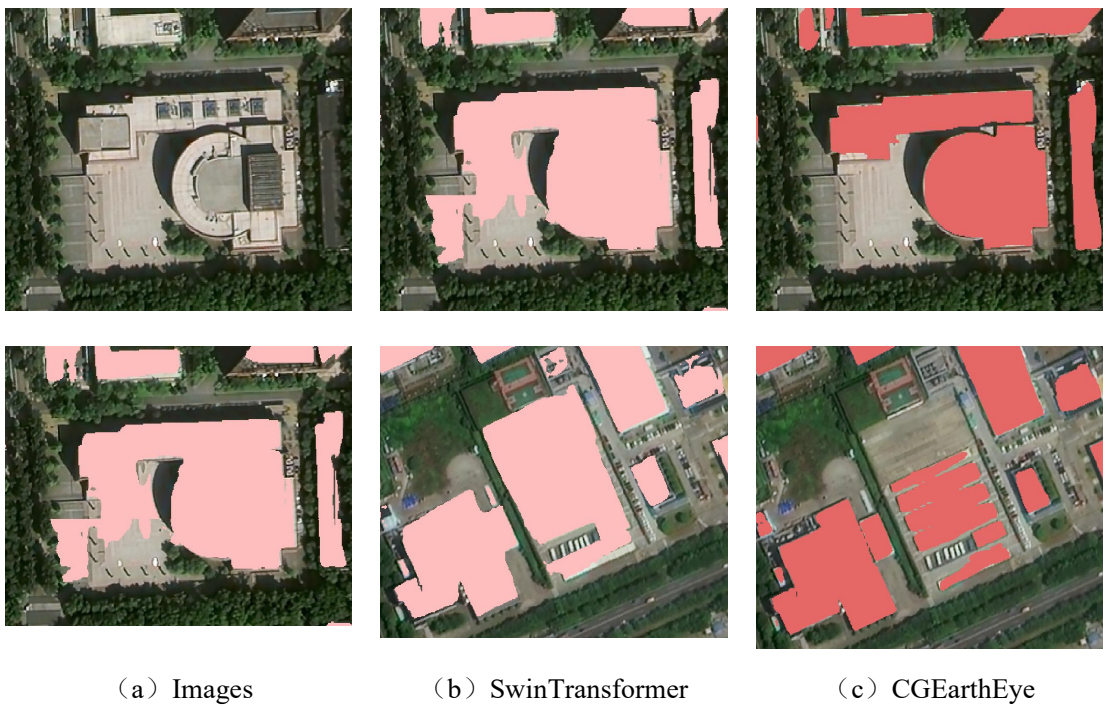

（a）Images　　　　　　　（b）SwinTransformer　　　　　　（c）CGEarthEye

**Fig.7 Comparative visualization of building extraction in Longhua District, Shenzhen**

3) Building Change Detection

Building change detection holds significant practical value for urban management, disaster response, and land monitoring. We implement a CGEarthEye-backed ChangeFormer model trained on the Jilin-1 Building Change Detection Dataset, with comparative evaluations against baseline ChangeFormer and BAN frameworks. The dataset comprises 33,359 sub-meter resolution image-label pairs covering five Chinese regions: Changsha (Hunan), Nan'an (Fujian), Hefei (Anhui), Changchun (Jilin), and Liaoyang (Liaoning), focusing primarily on building construction and demolition. Applied to 0.75m resolution imagery of Longhua District from the first to third quarter of 2023, accuracy was assessed via quadrat sampling. Results (Table 11) demonstrate CGEarthEye's superior performance. It achieves the highest F1-score exceeding alternatives by 8 to 17 percent, optimal precision averaging 4 to 16 percent higher than other models, and best recall outperforming counterparts by 10 to 19 percent. These outcomes confirm CGEarthEye's exceptional feature representation capacity for effectively suppressing false positives while maintaining outstanding detection reliability.

**Table 11 Comparison of building change extraction in Longhua District, Shenzhen**

| Model | Detections | Ground Truths | Recall | Precision | F1 |
|---|---|---|---|---|---|
| ChangeFormer | 446 | 979 | 0.4709 | 0.6973 | 0.5622 |
| BAN | 493 | 979 | 0.5557 | 0.8215 | 0.6629 |
| CGEarthEye-B | 584 | 979 | 0.6547 | 0.8613 | 0.7440 |

Visualization results are presented in Figure 8. Detailed inspection reveals that CGEarthEye merges minor unchanged areas within clustered change regions while preserving object segmentation boundaries, ultimately producing object-level patches with enhanced visual coherence. Notably, for persistent construction activities within development sites, the model consistently extracts entire changed parcels as unified entities.

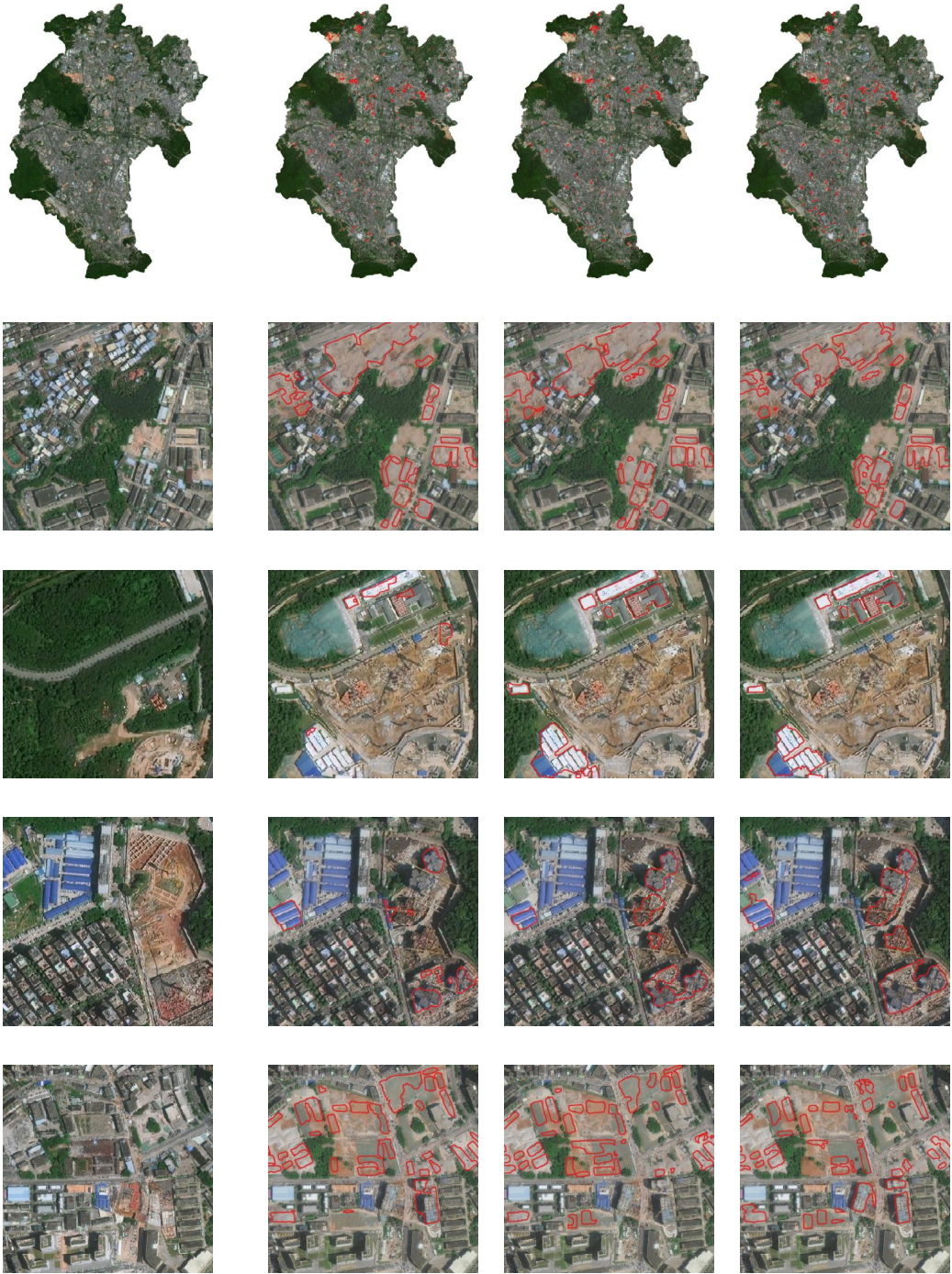

    (a)  Preliminary Imagery    (b)  ChangeFormer    (c)  BAN    (d)  CG-EarthEye

**Fig.8 Comparative visualization of building change extraction in Longhua District, Shenzhen.**

4.7 Efficiency Optimization

    Algorithmic efficiency becomes a critical constraint for regional-scale Earth observation applications when image resolution reaches meter or sub-meter levels, particularly as model parameters scale to billion-level magnitudes. To address computational limitations in practical deployments, we optimize CGEarthEye's downstream implementation framework. Fine-tuned downstream models leverage TensorRT deployment with

INT8 quantization and an optimized multithreaded I/O strategy for large-scale geospatial data. As benchmarked on a consumer-grade RTX 3090 GPU using 0.75m resolution imagery (Table 12), the optimized inference achieves $1.90\times$ and $2.31\times$ speedup over native PyTorch mixed-precision inference for CGEarthEye-Giant and CGEarthEye-Base respectively. Processing throughput reaches 5,536 km² /hour for CGEarthEye-Giant and 23,070 km² /hour for CGEarthEye-Base on single RTX 3090 GPU.

**Table 12 Inference speed on downstream tasks (square kilometers/hour)**

| Model | Semantic Segmentaion | | Object Detection | | Change Detection | |
|---|---|---|---|---|---|---|
| | native | optimized | native | optimized | native | optimized |
| CGEarthEye-B | 11000 | 24750 | 8640 | 23400 | 10600 | 21060 |
| CGEarthEye-G | 3300 | 5950 | 2890 | 5600 | 2550 | 5060 |

# 5  Conclusion

This study addresses the characteristics of the massive high-resolution satellite remote sensing data from Jilin-1 and proposes a high-resolution remote sensing visual foundation model framework, CGEarthEye. The framework includes a large-scale multi-temporal high-resolution dataset, a multi-granularity self-supervised learning strategy, and five ViT backbones with varying parameter scales, totaling 2.1 billion parameters. To enhance the representation performance of the foundation model, CGEarthEye employs multi-granularity self-supervised learning for pre-training on the world's first self-supervised dataset of over 15 million multi-temporal sub-meter-level images. In benchmark tests across 10 datasets covering four typical remote sensing observation tasks, the frozen-backbone CGEarthEye consistently achieves state-of-the-art (SOTA) performance. Additionally, CGEarthEye demonstrates robust image representation and generalization capabilities while being optimized for practical efficiency, making it highly effective in real-world Earth observation applications. In the future, this research will continue to expand multimodal datasets to enhance CGEarthEye's potential in multimodal data applications and better facilitate the synergistic use of Jilin-1 high-resolution data with other datasets. We believe that, with the integration of RSVFM models and satellite constellations, commercial aerospace will continue to drive deeper scientific advances in field of EO.